\documentclass[letterpaper]{article} 
\usepackage{aaai25}  
\usepackage{times}  
\usepackage{helvet}  
\usepackage{courier}  
\usepackage[hyphens]{url}  
\usepackage{graphicx} 
\usepackage{natbib}  
\usepackage{caption} 
\usepackage{algorithm}
\usepackage{algorithmic}
\usepackage{amsmath}
\usepackage{xspace}
\usepackage{multirow}

\newcommand{\model}{BiDeV\xspace}

\newcommand{\eg}{\emph{e.g.,}\xspace}

\usepackage{newfloat}
\usepackage{listings}
\DeclareCaptionStyle{ruled}{labelfont=normalfont,labelsep=colon,strut=off} 
\floatstyle{ruled}
\newfloat{listing}{tb}{lst}{}
\floatname{listing}{Listing}
\title{\model: Bilateral Defusing Verification for Complex Claim Fact-Checking}
\author{
    Yuxuan Liu\textsuperscript{\rm 1},
    Hongda Sun\textsuperscript{\rm 1},
    Wenya Guo\textsuperscript{\rm 2},
    Xinyan Xiao\textsuperscript{\rm 3},
    Cunli Mao\textsuperscript{\rm 4},
    Zhengtao Yu\textsuperscript{\rm 4},
    Rui Yan\textsuperscript{\rm 1}\thanks{Corresponding author: Rui Yan (ruiyan@ruc.edu.cn)}
}
\affiliations{
    \textsuperscript{\rm 1} Gaoling School of Artificial Intelligence, Renmin University of China \\
    \textsuperscript{\rm 2} Nankai University \\
    \textsuperscript{\rm 3} Baidu Inc. \\
    \textsuperscript{\rm 4} Kunming University of Science and Technology 

    \{yuxuanliu, sunhongda98, ruiyan\}@ruc.edu.cn, wenyaguo@nankai.edu.cn, \\ xiaoxinyan@baudi.com, maocunli@163.com, ztyu@hotmail.com
}

\usepackage{bibentry}

\begin{document}

\maketitle

\begin{abstract}
Complex claim fact-checking performs a crucial role in disinformation detection. 
However, existing fact-checking methods struggle with claim vagueness, specifically in effectively handling latent information and complex relations within claims.
Moreover, evidence redundancy, where nonessential information complicates the verification process, remains a significant issue.
To tackle these limitations, we propose \textit{\textbf{Bi}lateral \textbf{De}fusing \textbf{V}erification} (\textbf{\model}), a novel fact-checking working-flow framework integrating multiple role-played LLMs to mimic the human-expert fact-checking process. 
\model consists of two main modules: \textit{Vagueness Defusing} identifies latent information and resolves complex relations to simplify the claim, and \textit{Redundancy Defusing} eliminates redundant content to enhance the evidence quality.
Extensive experimental results on two widely used challenging fact-checking benchmarks (Hover and Feverous-s) demonstrate that our \model can achieve the best performance under both gold and open settings. This highlights the effectiveness of \model in handling complex claims and ensuring precise fact-checking\footnote{Code: \url{https://github.com/EthanLeo-LYX/BiDeV}}. 
\end{abstract}

%

\section{Introduction}
Fact-checking is crucial for claim verification by collecting relevant evidence and determining their veracity~\cite{guo2022survey}.
Disinformation, concealed within plenty of daily news and reports, threatens the cyber environment and social stability~\cite{liu2024tiny}. Given its critical role in combating disinformation, complex claim verification has attracted considerable attention from both academics and industry professionals~\cite{thorne2018automated, jiang2020hover, pan2023factchecking}.

\begin{figure}[]
\centering
\includegraphics[width=0.95\linewidth]{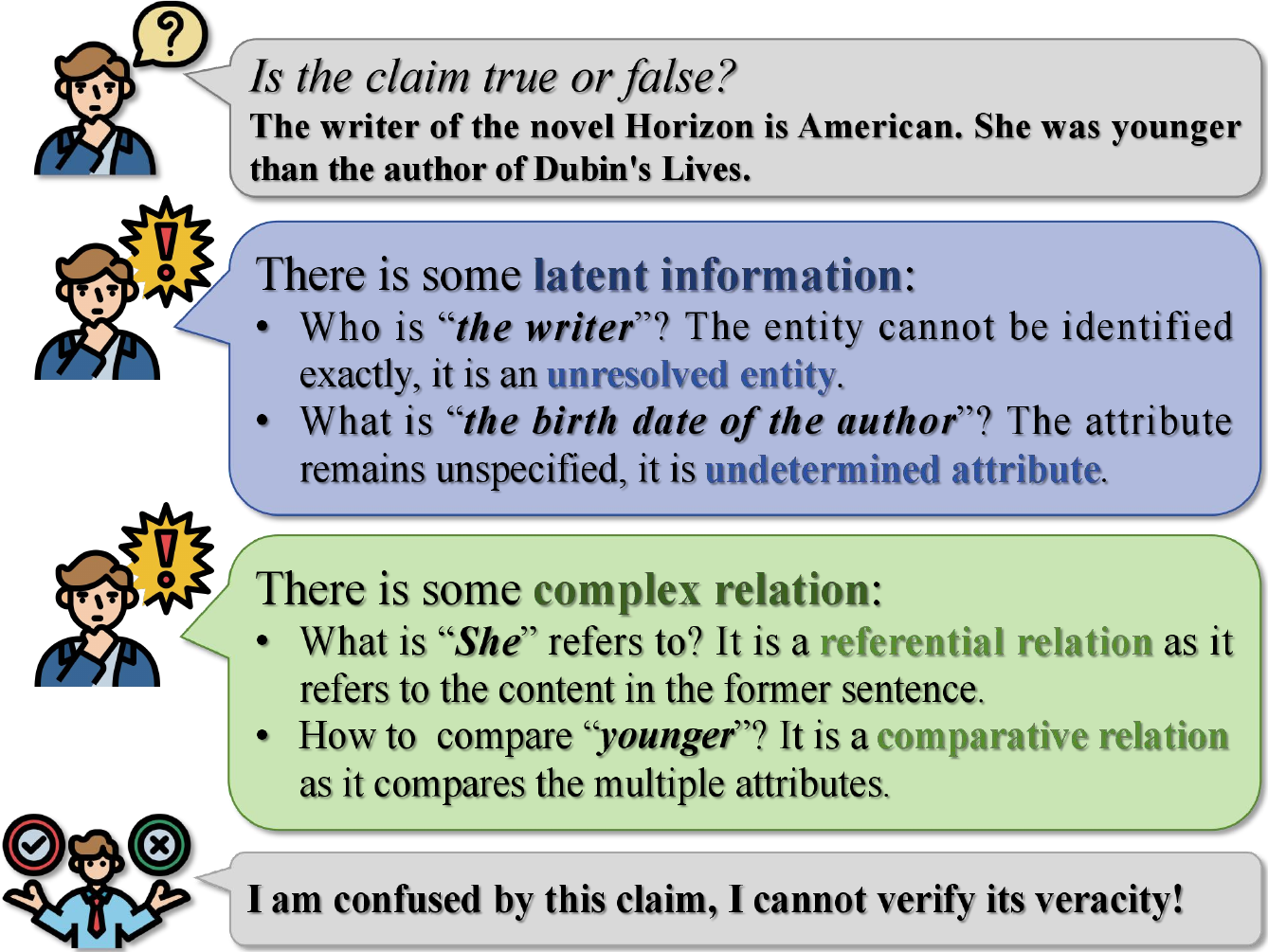}
\caption{An example illustrating how the claim vagueness impedes the fact-checking process. \textbf{Latent information} encompasses unresolved entities and undetermined attributes; \textbf{Complex relations} include referential relations and comparative relations.}
\label{fig:motivation}
\end{figure}

Recent fact-checking approaches can be broadly categorized into two categories: 
(i): \textit{Specialized Language Model (SLM)-based end-to-end methods} focus on extracting representations of claims and evidence then comparing them in the feature space for verification \cite{popat2018declare,soleimani2020bert}. Typically, they employ specific fine-tuned modules to establish correlations between claims and evidence~\cite{kruengkrai2021multi, xu2022evidence, liao2023muser}.
(ii): \textit{Large Language Model (LLM)-based step-by-step methods} leverage LLMs to conduct questioning or decomposing progressively~\cite{pan2023factchecking, zhang2023llmbased, wang2023explainable}.
These methods benefit from the advanced semantic understanding and reasoning capabilities of LLMs, enabling more nuanced and thorough fact-checking processes.

Despite some promising advancements, several challenges persist in current fact-checking methods, particularly concerning \textit{claim vagueness} and \textit{evidence redundancy}.
Claim vagueness poses a primary obstacle in the fact-checking process.
Figure~\ref{fig:motivation} illustrates an example of obstacles due to the claim vagueness.
In terms of content and correlation, claim vagueness can be categorized into two primary types:
(i) \textit{Latent information} encompasses unresolved entities that cannot be identified explicitly and undetermined attributes that remain unspecified.
(ii) \textit{Complex relations} include referential relations, where pronouns reference entities within the claim, and comparative relations, which compare multiple attributes.
Addressing these aspects is crucial for accurately clarifying claims, yet previous approaches often fall short in comprehensively handling these nuances, leading to inadequate verification performance~\cite{liao2023muser, rani2023factify}.
The quality of evidence is essential for claim verification. However, original documents often contain extensive, irrelevant details. This redundancy complicates fact-checking, as current methods overly rely on the evidence and fail to effectively filter out unnecessary information, leading to increased complexity and distraction during the fact-checking process~\cite{zou2023decker, zhang2023llmbased}.

To address these challenges, we aim to improve the complex claim fact-checking from two aspects: (i) \textit{claim simplification} identifies the latent information and resolves the complex relations to simplify the claim; (ii) \textit{evidence selection} retains the pertinent evidence and exclude the redundant content.
To this end, we propose \underline{\textbf{Bi}}lateral \underline{\textbf{De}}fusing \underline{\textbf{V}}erification (\textbf{\model}), a novel complex claim fact-checking working-flow framework that integrates multiple role-played LLMs to imitate the human-expert fact-checking process. 
To effectively tackle claim vagueness and evidence redundancy, \model incorporates two dedicated modules: (i) \textbf{Vagueness Defusing (VD)} formulates claim simplification into two stages: \textit{perceive-then-rewrite} iteratively identifies latent information in the claim, generates corresponding queries for explicit background information, and rewrites the claim for clarity; \textit{decompose-then-check} decomposes the simplified claim, resolves the complex relations, and verifies each sub-claim step by step; (ii) \textbf{Redundancy Defusing (RD)} evaluates and filters evidence based on the relevance to specific queries, thus obtaining more precise and pertinent evidence.
The VD module aims to simplify claims, reducing the complexity of the fact-checking process by eliminating vagueness.
Meanwhile, the RD module enhances the evidence quality by excluding irrelevant content, thus minimizing distractions during verification.

We conduct comprehensive experiments on widely used challenging complex claim fact-checking benchmarks: Hover \cite{jiang2020hover} and Feverous-s \cite{pan2023factchecking}. Experimental results
show that \model achieves the best performance, improving Macro-F1 by 3.88\% in both annotated evidence (gold) and retrieved evidence (open) settings. This indicates the effectiveness of the proposed VD and RD modules. 
Also, \model exhibits remarkable improvements on more complex claims, highlighting its competitive generalization ability in handling intricate scenarios.

Overall, our contributions can be summarized as follows:

$\bullet$ 
We propose \model, a novel fact-checking working-flow framework integrating LLMs to eliminate the vague information in the claim and the noisy redundancy in the evidence, which imitates the fact-checking process of the human experts.

$\bullet$
We introduce the vagueness defusing module formulated as a two-stage process fact-checking a complex claim through perceive-the-rewrite and decompose-then-check. This module concentrates on ascertaining latent information and resolving complex relations, contributing to reducing the complexity of fact-checking complex claims.

$\bullet$ 
We present the redundancy defusing module to filter out irrelevant information leading to more effective and pertinent evidence in sub-claim verification. Extensive experimental results demonstrate that \model greatly advances the performance in complex claim fact-checking.

\section{Related Work}

\textbf{Complex claim fact-checking} aims to identify factual conflicts existing between the claim and the given evidence, which serves as a pivotal technique to address fake news and rumor detection~\cite{liu2024skepticism}.

Previous works can be categorized as \textit{SLM-based end-to-end methods}, which focus on obtaining more effective representations of claims and evidences to conduct verification by comparing them in the feature space \cite{popat2018declare, ma2019sentence}. 
Utilizing specific models pre-trained or fine-tuned on some NLI datasets allows them to outperform traditional methods on fact-checking \cite{kruengkrai2021multi, he2022debertav3, wadden2022multivers}. Moreover, designing some specific modules to correlate the claim and evidence is necessary to achieve more precise verification \cite{xu2022evidence, liao2023muser}. 

\begin{figure*}
\centering
\includegraphics[width=0.99\linewidth]{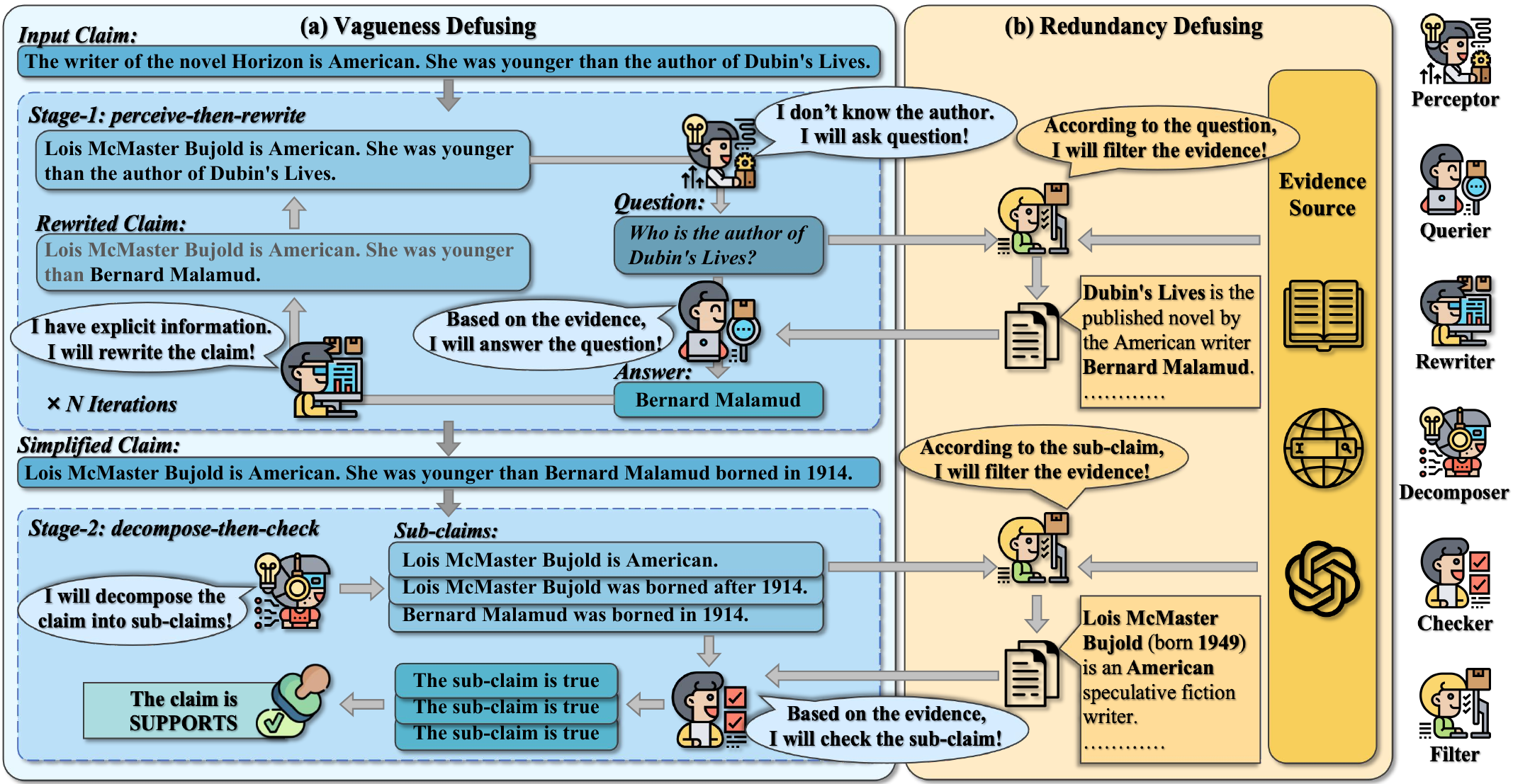}
\caption{The overview of our \model. Two main modules for Bilateral Defusing Verification: (a) \textbf{Vagueness Defusing} for input claim. \textit{Perceive-then-rewrite} stage simplifies the claim iteratively: the perceptor perceives questions about latent information, the querier provides explicit knowledge to the question and the rewriter rewrites the latent information in the claim with the explicit knowledge. \textit{Decompose-then-check} stage verifies the claim: the decomposer splits several sub-claims and the checker verifies the sub-claims. (b) \textbf{Redundancy Defusing} for evidence. The evidence extracted from the source is refined by the filter.}
\label{fig:model}
\end{figure*}

As large language models (LLMs) have demonstrated advances in reasoning~\cite{wei2022chain,wang2022self,sun2024determlr}, various LLM roles have achieved success in different fields~\cite{sun2024harnessing,sun2024facilitating,liu2025mobile}.
Recent works instruct LLMs to think step-by-step to gradually conduct fact-checking, such as iteratively questioning \cite{press2022measuring} and program-guided reasoning \cite{pan2023factchecking}. Some approaches split the complex claim into several simple sub-claims, which reduce the difficulty of verifying each sub-claim \cite{zhang2023llmbased, wang2023explainable}.

However, previous works have not adequately addressed vague information in the claim and noisy redundancy in the evidence, which limits their performance. To address these issues, we propose \model, which imitates the verification process of human experts, to achieve accurate complex claim fact-checking through more effective claim simplification and evidence selection.

\section{Bilateral Defusing Verification}

\subsection{Task Formulation}

The complex claim fact-checking task places a central emphasis on verifying the veracity of the claim based on the pertinent evidence. Specifically, given a claim $C$, an evidence source $S$, a fact-checking model $M$ concentrates on predicting the veracity label $Y$ using the evidence from $S$. 
\begin{equation}
    Y = M(C, S),\quad Y\in[\text{Support}, \text{Refute}]
\end{equation}

We address that complex claim fact-checking focuses on claim simplification and evidence selection, thus we formulate our working-flow framework as two main modules: \textbf{Vagueness Defusing} for claims and \textbf{Redundancy Defusing} for evidence. The overview of our \model is shown in Figure \ref{fig:model}. In the subsequent sections, we will introduce how to integrate LLMs to eliminate the vagueness in the claim and the redundancy in the evidence.

\subsection{Vagueness Defusing}

As shown in Figure \ref{fig:motivation}, complex claims often contain two types of vagueness: latent information and complex relations. These elements increase the complexity of fact-checking. When human experts face such complex claims, they first query for undetermined information to obtain explicit background knowledge. Then, they analyze and reconstruct the claim based on the collected background knowledge to eliminate this undetermined information, unravel the complex internal correlations to split several sub-claims, and finally verify the sub-claims to derive the ultimate result \cite{nakov2021automated, das2023state, pan2023factchecking}. 
To imitate the iterative process of human experts, we divide the vagueness defusing process into two stages: \textit{perceive-then-rewrite} for latent information and \textit{decompose-then-check} for complex relations.

\noindent \textbf{Stage-1: Perceive-then-Rewrite.} 
Latent information can be classified into two categories: unresolved entities and undetermined attributes. In the example shown in Figure \ref{fig:motivation}, \textit{``the writer''} is an unresolved entity since its reference is not specified within the claim; \textit{``the birth date''} is an undetermined attribute as it is not mentioned in the claim.
To defuse the latent information, we instruct LLMs to implement an iterative and collaborative process involving three roles: the perceptor, querier, and rewriter. This process transforms the initial complex claim $C$ into the simplified claim $C^*$ step by step. The details of their working order are discussed below.

\noindent $\bullet$ \textbf{Perceptor ($M_p$)}
is performed by the LLM through a step-by-step thinking process to perceive latent information in the following standard: 
(1) An entity is considered to be unresolved if the entity it refers to cannot be found in the claim; 
(2) An attribute is considered to be undetermined if the attribute of the subject is not mentioned in the claim. 
Specifically, given the rewritten claim $c_{i-1}$ in the $(i-1)_{th}$ iteration, the perceptor is responsible for accurately identifying both types of latent information and generating targeted question $q_i$ for explicit background knowledge:
\begin{equation}
    q_i = M_p(c_{i-1})
\end{equation}

\noindent $\bullet$ \textbf{Querier ($M_q$)}
is responsible for answering the question generated by the perceptor for precise and explicit content of latent information.
Given the question $q_i$ generated by $M_p$, we instruct the LLM to comprehend and integrate pertinent information within the evidence $e^*_i$ extracted and refined from evidence source $S$, then generate a precise and dependable answer $a_i$: 

\begin{equation}
    a_i = M_q(q_i, e^*_i)
\end{equation}

\noindent $\bullet$ \textbf{Rewriter ($M_r$)} is essential at this stage as it integrates explicit background knowledge and simplifies the statement of the claim.
Since the claim may contain complex internal correlations, 
merely using these QA pairs as supplementary evidence is insufficient for verification.
Given the question $q_i$ in the $i_{th}$ iteration, the rewriter first finds the direct counterparts and the indirect relevance in the claim $c_{i-1}$, then rewrites them using the answer $a_i$. The rewriting process can be formulated as:

\begin{equation}
    c_i = M_r(c_{i-1}, q_i, a_i)
\end{equation}

\noindent \textbf{Stage-2: Decompose-then-Check.}
After the perceive-then-rewrite stage, the simplified claim $C^*$ has effectively reduced the latent information but may still contain some complex relations: referential relation and comparative relation. As shown in Figure \ref{fig:motivation}, \textit{``She''} is a referential relation as it refers to \textit{``the writer''} in the former sentence; \textit{``younger''} is a comparative relation as it compares the birth date of \textit{``She''} and \textit{``the author''}. 
To further clarify claims, we employ a decomposer to disentangle these complex relations and a checker to perform more detailed verification.

\noindent $\bullet$ \textbf{Decomposer ($M_d$)} 
is performed by the LLM to resolve the complex relations: it replaces referential relations with explicit entities in the claim and splits comparative relations using determined attributes. Then the complex claim is decomposed into a series of brief declarative sub-claims with simple logic and unitary content. 
Given the simplified claim $C^*$, the process of decomposing sub-claims $sc$ is given by:
\begin{equation}
    sc = M_d(C^*)
\end{equation}

\noindent $\bullet$ \textbf{Checker ($M_c$)}
conducts the final step of fact-checking to verify each sub-claims and conclude the veracity result of the entire claim. Since the claim and evidence may describe the same facts in different ways, we guide the LLM to comprehensively understand and extract valuable insights from evidence, then integrate and match them with the claim, and finally produce a dependable result after meticulous reasoning.
With the relevant evidence $e^*_j$, we obtain the verification result $y_j$ of the sub-claims $sc_j$, and ultimately conclude the predicted veracity label $Y$ of the entire claim:
\begin{equation}
    Y =\bigcap_j^{|sc|} y_j, \quad y_j = M_c(sc_j, e^*_j)
\end{equation}

\subsection{Redundancy Defusing}

\begin{table*}[]
\renewcommand\arraystretch{1.35}
\setlength{\tabcolsep}{5.5pt}
\centering
\begin{tabular}{l|cc|cc|cc|cc}
\hline \hline
\multirow{2}{*}{\textbf{Methods}} &
  \multicolumn{2}{c|}{\textbf{Hover(hop2)}} &
  \multicolumn{2}{c|}{\textbf{Hover(hop3)}} &
  \multicolumn{2}{c|}{\textbf{Hover(hop4)}} &
  \multicolumn{2}{c}{\textbf{Feverous-s}} \\ \cline{2-9} 
 &
  \textbf{Gold} &
  \textbf{Open} &
  \textbf{Gold} &
  \textbf{Open} &
  \textbf{Gold} &
  \textbf{Open} &
  \textbf{Gold} &
  \textbf{Open} \\ \hline
Bert-FC$^*$\space\cite{soleimani2020bert}       & 53.41 & 50.68 & 50.91 & 49.86 & 50.86 & 48.57 & 74.71 & 51.67 \\
LisT5$^*$\space\cite{jiang2021exploring}         & 56.15 & 52.56 & 53.76 & 51.89 & 51.67 & 50.46 & 77.88 & 54.15 \\ \hline
RoBERTa-NLI$^*$\space\cite{nie-etal-2020-adversarial}   & 74.62 & 63.62 & 62.23 & 53.99 & 57.98 & 52.41 & 88.28 & 57.81 \\
DeBERTaV3-NLI$^*$\space\cite{he2022debertav3} & 77.22 & 68.72 & 65.98 & 60.76 & 60.49 & 56.01 & 91.98 & 58.81 \\
MULTIVERS$^*$\space\cite{wadden2022multivers}     & 68.86 & 60.17 & 59.87 & 52.55 & 55.67 & 51.86 & 86.03 & 56.61 \\ \hline
Codex$^*$\space\cite{chen2021evaluating}         & 70.63 & 65.07 & 66.46 & 56.63 & 63.49 & 57.27 & 89.77 & 62.58 \\
FLAN-T5$^*$\space\cite{chung2022scaling}       & 73.69 & 69.02 & 65.66 & 60.23 & 58.08 & 55.42 & 90.81 & 63.73 \\ \hline
HiSS$^\dag$\space\cite{zhang2023llmbased}          & 73.06 & 66.25 & 65.14 & 58.56 & 64.67 & 57.64 & 89.26 & 65.99 \\
FOLK$^\dag$\space\cite{wang2023explainable}          & 73.24 & 67.29 & 65.84 & 58.61 & 64.73 & 58.79 & 89.52 & 66.89 \\
ProgramFC$^\dag$\space\cite{pan2023factchecking}     & 74.59 & 69.89 & 66.75 & 61.21 & 65.00 & 58.21 & 91.23 & 67.22 \\ 
Factcheck-GPT$^\dag$\space\cite{wang2023factcheck}   & 74.88 & 70.25 & 66.32 & 60.11 & 66.62 & 59.25 & 91.39 & 67.24 \\
\hline
\model &
  \textbf{77.59} &
  \textbf{73.44} &
  \textbf{69.91} &
  \textbf{63.62} &
  \textbf{70.63} &
  \textbf{60.41} &
  \textbf{92.39} &
  \textbf{69.01} \\ \hline \hline
\end{tabular}
\caption{Macro-F1 scores of \model and baselines on Hover and Feverous-s under both gold and open settings. Compared baselines include: (i) \textit{Pre-trained methods}; (ii) \textit{Fine-tuned methods}; (iii) \textit{LLM-ICL methods}; and (iv) \textit{LLM-reason methods}. Bold numbers indicate significant improvements ($p<0.05$) based on 10 rounds of bootstrapping sampling. Results with $^*$ are quoted from~\cite{pan2023factchecking}, and results with $^\dag$ are reproduced by \texttt{gpt-3.5-turbo} for a fair comparison.}
\label{tab:main}
\end{table*}

When answering questions and verifying claims, human experts first extract potential evidence and then select pertinent paragraphs providing precise and credible information. Hence, we emulate this process by initially extracting coarse-grained relevant evidence from the evidence source $S$. For the gold setting, we directly use the evidence from annotated with the gold labels in the dataset. For the open setting, we retrieve evidence from external knowledge bases (\eg Wikipedia). 
However, the initially extracted evidence often contains redundant and noisy information, which can confuse the querier and checker. Therefore, we filter out the irrelevant information through step-by-step thinking.

\noindent $\bullet$ \textbf{Filter ($M_f$)} 
firstly segments the initially extracted evidence into multiple paragraphs and then evaluates whether each paragraph is relevant to the question or the sub-claim, which involves not only directly relevant content but also potentially contributed information. The irrelevant paragraphs are eliminated to get the most imperative and effective evidence.
As an example of answering a question $q_i$, given the extracted evidence $e_i$, we obtain the filtered evidence $e^*_i$ by:
\begin{equation}
    e^*_i = M_f(e_i, q_i)
\end{equation}

\section{Experiments}

\subsection{Experimental Setup}

\noindent \textbf{Datasets.}
There are two widely used and challenging datasets to evaluate the fact-checking performance of baselines and our \model: (i) \textbf{Hover} \cite{jiang2020hover} and (ii) \textbf{Feverous-s} \cite{pan2023factchecking}.
Both of the datasets need to verify the given claim with multiple evidences through multi-step reasoning. 

\noindent \textbf{Baselines.}
To demonstrate the effectiveness of our method, we compare
\model with the following four types of baselines:
(i) \textit{Pre-trained methods}: 
BERT-FC \cite{soleimani2020bert} and LisT5 \cite{jiang2021exploring}.
(ii) \textit{Fine-tuned methods}: 
RoBERTa-NLI \cite{nie-etal-2020-adversarial}, DeBERTaV3-NLI \cite{he2022debertav3} and MULTIVERS \cite{wadden2022multivers}.
(iii) \textit{LLM-ICL methods}: 
FLAN-T5 \cite{chung2022scaling} and Codex \cite{chen2021evaluating}.
(iv) \textit{LLM-reason methods}: 
Hiss \cite{zhang2023llmbased}, FOLK \cite{wang2023explainable}, ProgramFC \cite{pan2023factchecking} and FactcheckGPT \cite{wang2023factcheck}.

\noindent \textbf{Evaluation Metrics.}
We use Macro-F1 as metrics in order to better deal with unbalanced proportions between \textit{support} and \textit{refute} samples.

\noindent \textbf{Implementation Details.}
In our proposed method, we use \texttt{gpt-3.5-turbo} as the base model of Perceptor, Rewriter, Decomposer, and Filter by accessing to OpenAI API with few-shot demonstrations.
For a fair comparison, we leverage Flan-T5-XL (3B) as the Querier and Checker without additional fine-tuning.
In the vagueness defusing, we iteratively perceive-then-rewrite for 3 rounds.
To evaluate in the open setting, we use BM25 \cite{robertson2009probabilistic} to retrieve top-K (K=10) evidence documents. 

\subsection{Overall Performance \label{sec:overall_performance}}

We evaluate \model and the compared baselines on two challenging benchmarks under two settings: annotated evidence as gold-setting and retrieved evidence as open-setting. The overall performance is shown in Table \ref{tab:main}. The experimental results demonstrate the following conclusions.

$\bullet$ \noindent \textit{\model achieves the best performance.}
Our \model achieves appealing performance improvement against 11 baselines from 4 categories. 
Specifically, \model outperforms fine-tuned baselines by 10.69\% (gold) and 15.27\% (open) on average without training.
Compared to both LLM-based baselines, \model also obtains 6.22\% performance improvement. 
The experiment results demonstrate that our proposed \model could achieve outstanding performance gains.

$\bullet$ \noindent \textit{\model improves on more complex claims.}
Although DeBERTaV3-NLI could be competitive with \model on easier 2-hop claims, its performance drops extremely as the complexity increases, and \model surpasses it by 5.33\% and 12.31\% on 3-hop and 4-hop claims. 
Overall, \model achieves improvement by 10.86\%@2-hop, 11.72\%@3-hop, and 17.72\%@4-hop under gold-setting, which indicates that \model performs more effectively on complex claims.

$\bullet$ \noindent \textit{Integrating perceiving, rewriting, and decomposing is effective.}
Compared with decomposition-based Hiss, question-based FOLK, and program-guided ProgramFC, 
\model surpasses them by 5.67\% on average, which demonstrates that the integration of perceiving, rewriting, and decomposing could inject explicit background information, resolve intricate correlations, simplify the claim, and reduce the complexity of fact-checking.

\begin{figure}
\centering
\includegraphics[width=\linewidth]{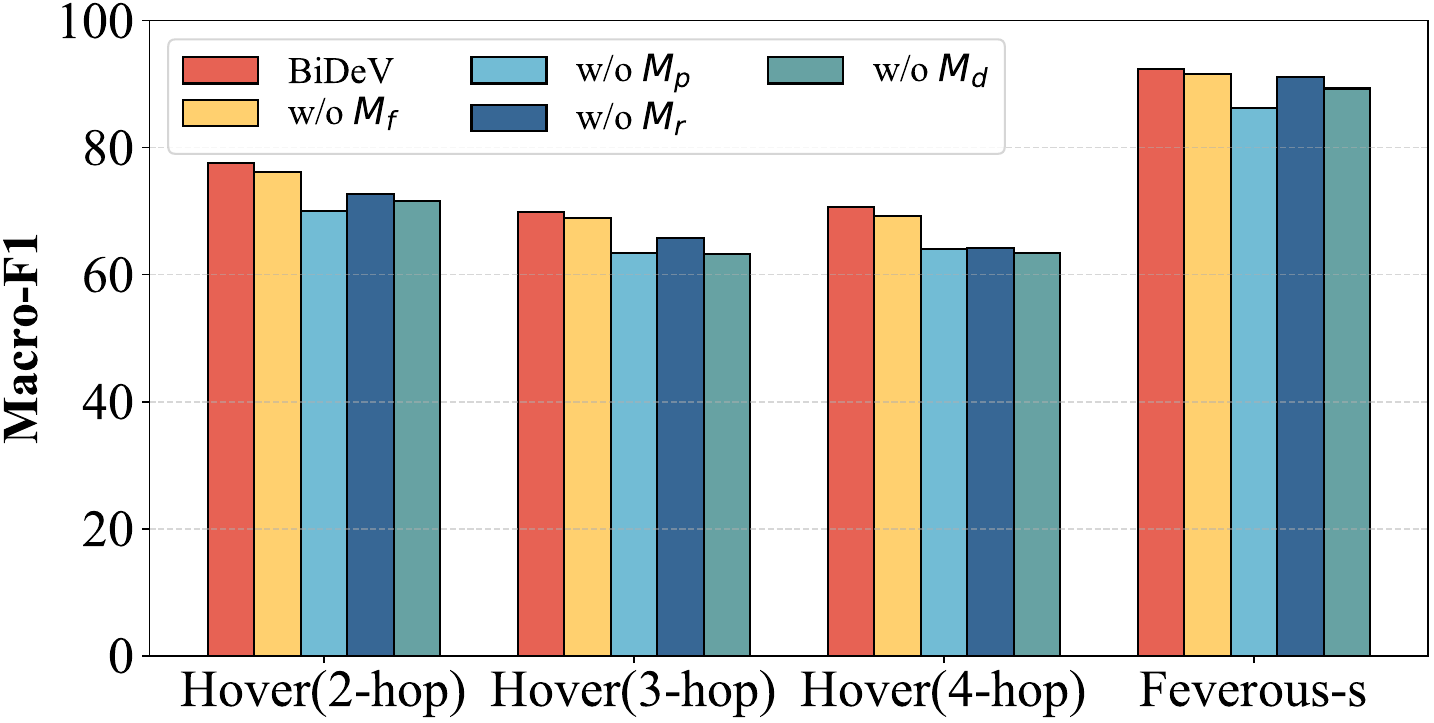}
\caption{Ablation study on Hover and Feverous-S. $M_f$: Filter; $M_p$: Perceptor; $M_r$: Rewriter; $M_d$: Decomposer.}
\label{fig:ablation}
\end{figure}

\subsection{Ablation Study}

In this section, we eliminate perceptor, rewriter, decomposer, and filter respectively, and explore to what extent these modules have an impact on the complex claim fact-checking. 
We conducted an ablation study on the gold setting, which is more representative because of its balanced performance. 
As shown in Figure \ref{fig:ablation}, the perceptor has the most impact, which indicates that generating questions for explicit background information is necessary. The decomposer contributes to the verification as it disentangles a complex claim into several brief sub-claims that are much easier to be verified. The rewriter could reduce the complexity of understanding claims as well. These three modules demonstrate that vagueness defusing is effective in simplifying the claim and leading to better fact-checking accuracy. The feasibility of the filter has also confirmed that redundancy defusing could estimate and improve the evidence quality.

\begin{figure}
\centering
\includegraphics[width=0.94\linewidth]{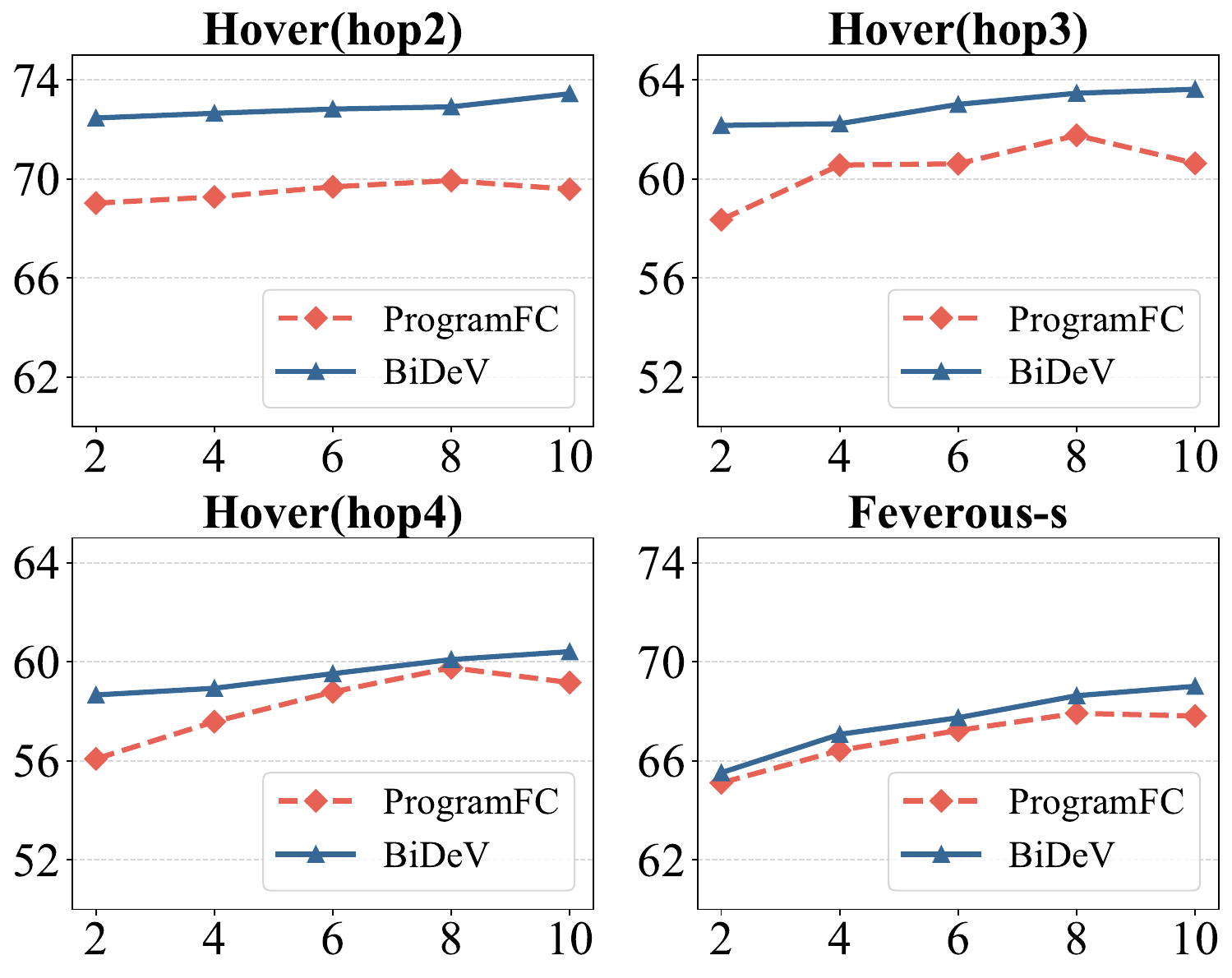}
\caption{Analysis of Redundancy Defusing under different Top-K retrieved evidence.}
\label{fig:retrieve_n}
\end{figure}

\subsection{Additional Analysis \label{sec:analysis}}

\begin{figure*}
\centering
\includegraphics[width=0.9\linewidth]{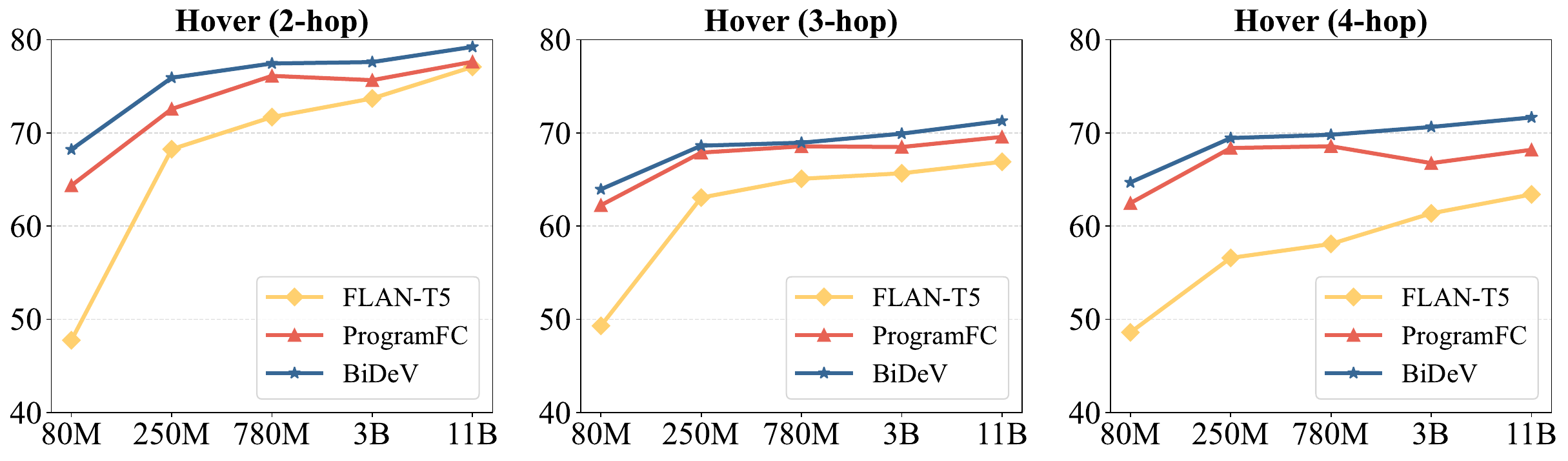}
\caption{Analysis of different model scales in Querier and Checker: FLAN-T5-small (80M), FLAN-T5-base (250M), FLAN-T5-large (780M), FLAN-T5-XL (3B), FLAN-T5-XXL (11B) on Hover 2-hop, 3-hop, and 4-hop subsets.}
\label{fig:model-scale}
\end{figure*}

\noindent \textbf{Analysis of Redundancy Defusing.}
We conducted comparative experiments with a selected strong baseline ProgramFC to explore the performance in retrieving different numbers of evidence and the results are shown in Figure \ref{fig:retrieve_n}.
Intuitively, more evidence will provide more information, leading to more accurate fact-checking. However, the performance of ProgramFC exhibits an upward-then-downward trend as the number increases, because the gain from useful information is offset by the interference from redundant information when too many evidences are retrieved.
Compared to ProgramFC, \model achieves consistent performance improvement as the number of retrieved evidences increases, which demonstrates that redundancy defusing module performs fine-grained filtering from the extracted evidences to obtain more pertinent and effective information.

\begin{figure}
\centering
\includegraphics[width=0.95\linewidth]{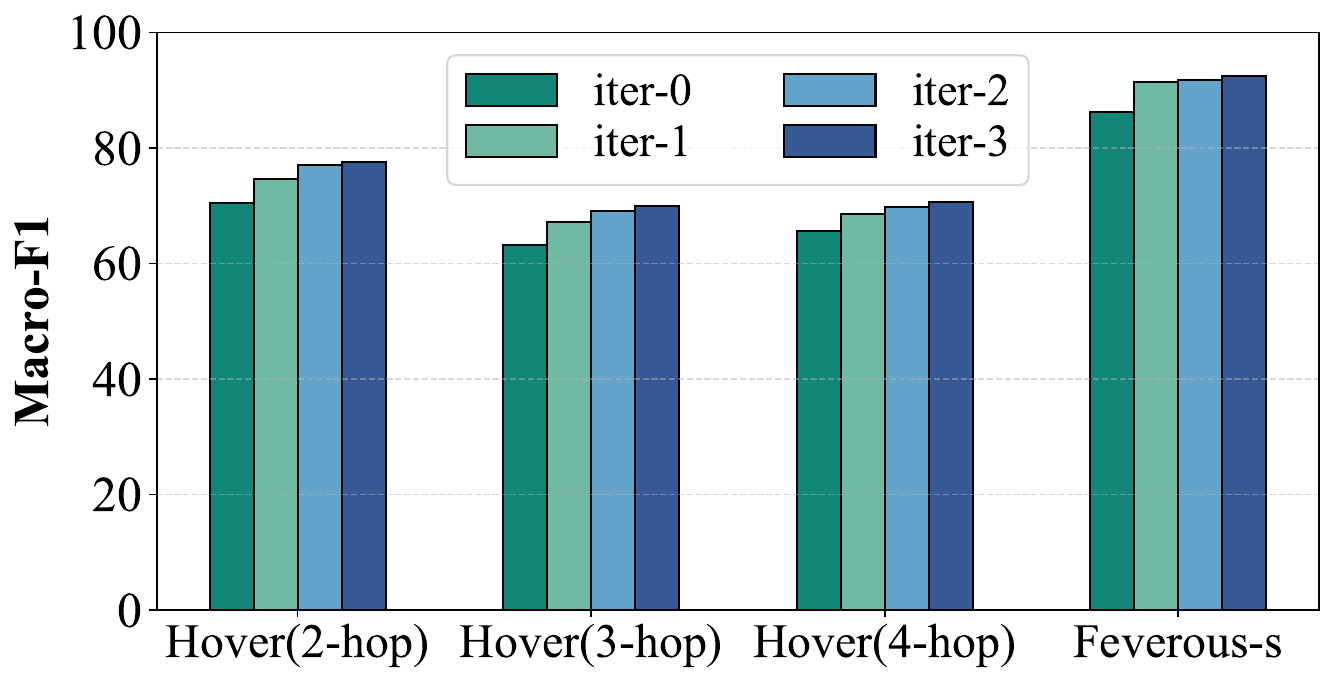}
\caption{Analysis of different iteration numbers of \textit{perceive-then-rewrite} in Vagueness Defusing.}
\label{fig:iteration_num}
\end{figure}

\noindent \textbf{Analysis of Vagueness Defusing.}

\noindent $\bullet$ \textbf{Iteration of perceive-then-rewrite.}
\textit{Perceive-then-rewrite} is an iterative process designed to involve more precise information and simplify the complex claim.
We designed experiments to investigate the effect of different numbers of iterations on the verification accuracy. 
As shown in Figure \ref{fig:iteration_num}, with the increase in the number of iterations, the accuracy of fact-checking gradually increases and tends to stabilize. The experimental results reveal that it is necessary and effective to constantly rewrite the claim based on queried explicit background knowledge, which eliminates vague information and simplifies the claim.
In the trade-off between performance and cost, we finally set the maximum number of iterations to 3 according to the results.

\noindent $\bullet$ \textbf{Strategies of decomposition.}
Decomposition plays an important role in \textit{decompose-then-verify}, thus we explore the effects of different decomposition strategies: 
(1) Direct: directly verify the simplified claim; 
(2) Naive: naively decompose the simplified claim; 
(3) \model: decompose the simplified claim to resolve complex relations.
We conduct evaluation on both gold and open settings, the experimental result is shown in Table \ref{tab:decompose-strategy}. Comparison with Direct demonstrates the necessity of the decomposition, and comparison with Naive proves the effectiveness of the complex relation-oriented decomposition in \model.

\begin{table}[]
\renewcommand\arraystretch{1.3}
\centering
\begin{tabular}{lcccc}
\hline \hline
\multicolumn{1}{l|}{\multirow{2}{*}{\textbf{Strategy}}} & \multicolumn{3}{c}{\textbf{Hover}}               & \multicolumn{1}{l}{\multirow{2}{*}{\textbf{Feverous-s}}} \\
\multicolumn{1}{l|}{}                          & \textbf{2-hop} & \textbf{3-hop} & \textbf{4-hop} & \multicolumn{1}{l}{}                            \\ \hline
\multicolumn{1}{l|}{Direct}           & 71.55          & 63.23          & 63.36          & 89.31                                           \\
\multicolumn{1}{l|}{Naive}           & 73.61          & 65.34          & 65.57          & 90.15                                           \\
\multicolumn{1}{l|}{\model}           & \textbf{77.59}          & \textbf{69.91}          & \textbf{70.63}          & \textbf{92.39}                                           \\ \hline
\multicolumn{1}{l|}{Direct}           & 66.26          & 59.23          & 57.17          & 60.28                                           \\
\multicolumn{1}{l|}{Naive}           & 66.85          & 60.05          & 57.65          & 65.41                                           \\
\multicolumn{1}{l|}{\model}           & \textbf{73.44}          & \textbf{63.62}          & \textbf{60.41}          & \textbf{69.01}                                           \\ \hline \hline
\end{tabular}
\caption{Analysis of different decomposition strategies. Above is the gold setting; Below is the open setting.}
\label{tab:decompose-strategy}
\end{table}

\begin{figure*}
\centering
\includegraphics[width=0.97\textwidth]{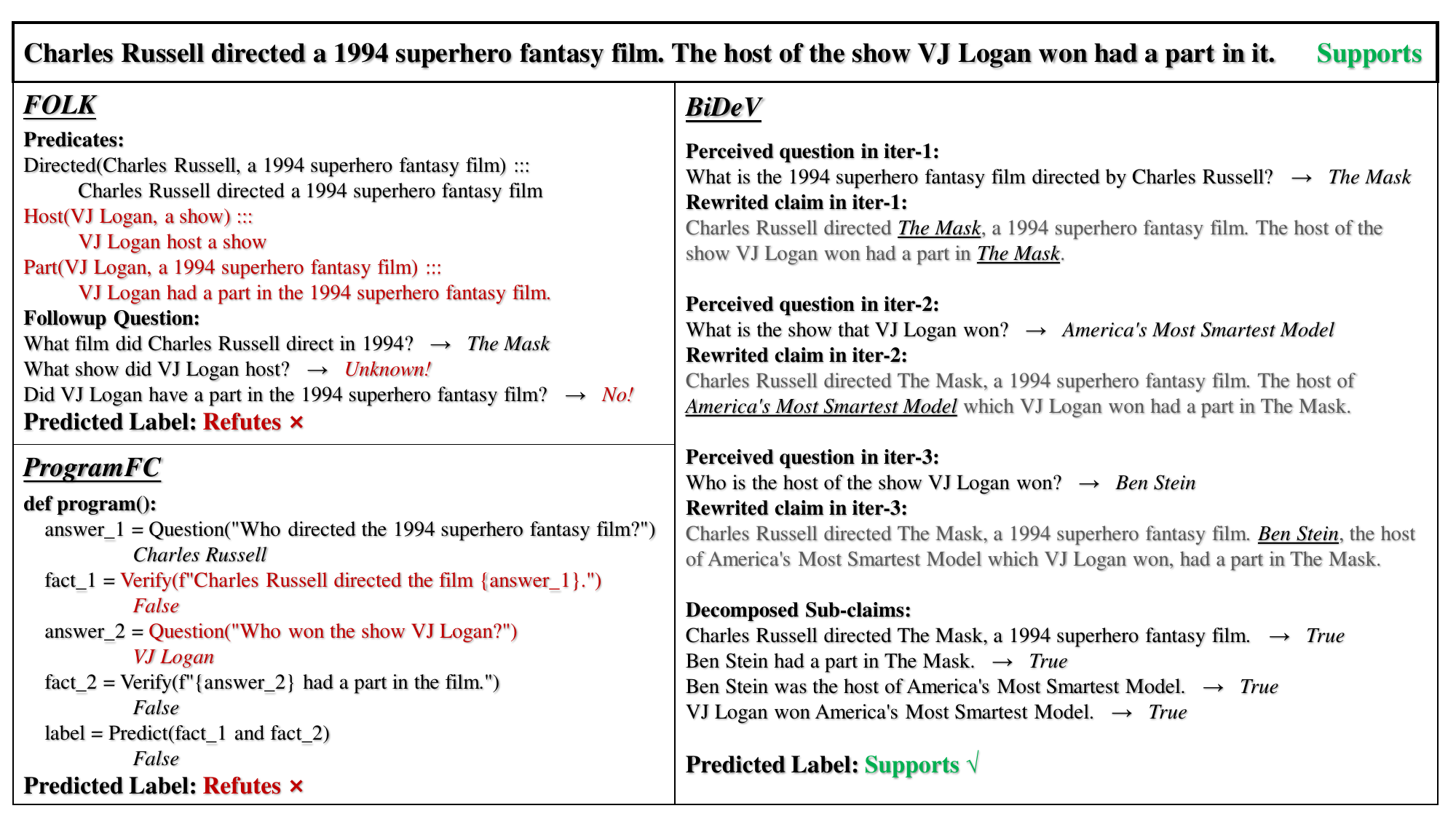}
\caption{Case Study of selected baselines (FOLK and ProgramFC) and our \model.}
\label{fig:case-study}
\end{figure*}

\noindent \textbf{Analysis of Querier and Checker.}
In our proposed \model, the accuracy of answering the questions affects the effectiveness of claim rewriting, and the verification of the sub-claim directly influences the overall fact-checking accuracy. Consequently, we scale the base model of Querier and Checker and conduct a comparison on Hover that is more direct to evaluate performance on different complexity of claims \cite{pan2023factchecking}. 
As shown in Figure \ref{fig:model-scale}, our \model allows for better generalization on larger-scale base models. Compared to FLAN-T5, the improvement is more on a smaller base model, 35.22\% on 80M parameters, because the reasoning ability is constrained by the model scale. Our bilateral defusing effectively alleviates this issue by simplifying the claim and selecting pertinent evidence. It reveals that \model better eliminates the obstacles to verifying complex claims that we surpass ProgramFC with the sub-task solver of 11B by only using the base model of 250M as Querier and Checker.

\noindent \textbf{Analysis of complex claim comprehension.}
We also conducted a close-setting experiment
that \textit{no} evidence is available and the model can only achieve better performance by comprehending and simplifying the claim more effectively.
The experimental results are shown in Table \ref{tab:close}. We surpass FLAN-T5 by 9.62\% on average, which indicates that our simplified claim can be checked effectively even by the limited knowledge stored in the parameters of the 3B model. 
Compared to different reasoning prompt methods of LLM, our \model achieves 5.49\% improvement on average. This demonstrates that vagueness defusing contributes to reducing the complexity of fact-checking. 
Moreover, \model also gains more improvement on more complex claims: 5.76\% on 2-hop and 6.39\% on 4-hop, which proves that our vagueness defusing module is more effective on complex claims. 

\subsection{Case Study}

\begin{table}[]
\renewcommand\arraystretch{1.3}
\centering
\begin{tabular}{l|cccc}
\hline \hline
\multirow{2}{*}{\textbf{Methods}} & \multicolumn{3}{c}{\textbf{Hover}}               & \multirow{2}{*}{\textbf{Feverous-s}} \\
          & \textbf{2-hop} & \textbf{3-hop} & \textbf{4-hop} &       \\ \hline
FLAN-T5$^*$    & 48.27          & 52.11          & 51.13          & 55.16 \\
Direct$^*$    & 56.51          & 51.75          & 49.68          & 60.13 \\
CoT$^*$       & 50.31          & 52.32          & 51.58          & 54.78 \\
Self-Ask$^*$  & 51.54          & 51.47          & 52.45          & 56.82 \\
ProgranFC$^\dag$ & 54.43          & 54.23          & 52.74          & 59.69 \\ \hline
\model                              & \textbf{56.73} & \textbf{54.82} & \textbf{53.66} & \textbf{61.12}                     \\ \hline \hline
\end{tabular}
\caption{Analysis of complex claim fact-checking under close-setting. Results with $^*$ are quoted from~\cite{pan2023factchecking}; Results with $^\dag$ are reproduced by \texttt{gpt-3.5-turbo}.}
\label{tab:close}
\end{table}

To present a more intuitive presentation of \model in the fact-checking process, we select FOLK and ProgramFC for comparison. As shown in Figure \ref{fig:case-study}, the vague information in the claim has been eliminated after \textit{perceive-then-rewrite} stage and the decomposed sub-claims have been verified successfully. However, FOLK generates invalid predicates leading to improper answers to the follow-up questions confused by the complex statement in the claim. Similarly, ProgramFC encounters wrong variable correlation and sub-task function calls. 
Both FOLK and ProgramFC are close to machine-centric reasoning, which is constrained by complex claims. In contrast, \model imitates the thinking process of human experts achieving more accurate fact-checking.

\section{Conclusion}

We propose Bilateral Defusing Verification (\model) in this paper, a novel framework integrating multiple LLMs to effectively imitate the complex claim fact-checking process of human experts. 
The vagueness defusing module eliminates latent information and resolves complex correlations, thereby simplifying the claims.
The redundancy defusing module filters out irrelevant evidence to provide more pertinent information for verification.
Experimental results show that \model advances the best performance on two challenging benchmarks (Hover and Feverous-s). This highlights \model's significant improvements in handling complex claims and offering more intuitive reasoning processes.

\section{Acknowledgements}
This work was supported by the National Natural Science Foundation of China (NSFC Grant No. 62122089 and 62302243),
Beijing Outstanding Young Scientist Program NO. BJJWZYJH012019100020098, and Intelligent Social Governance Platform, Major Innovation \& Planning Interdisciplinary Platform for the “Double-First Class” Initiative, Renmin University of China, the Fundamental Research Funds for the Central Universities, the Research Funds of Renmin University of China, and the Research Fund of Xiaomi.

\bibliography{aaai25}

\end{document}